# Residual Belief Propagation:
# Informed Scheduling for Asynchronous Message Passing

**Gal Elidan**
Computer Science Dept.
Stanford University
galel@cs.stanford.edu

**Ian McGraw**
Computer Science Dept.
Stanford University
imcgraw@cs.stanford.edu

**Daphne Koller**
Computer Science Dept.
Stanford University
koller@cs.stanford.edu

## Abstract

Inference for probabilistic graphical models is still very much a practical challenge in large domains. The commonly used and effective belief propagation (BP) algorithm and its generalizations often do not converge when applied to hard, real-life inference tasks. While it is widely recognized that the scheduling of messages in these algorithms may have significant consequences, this issue remains largely unexplored. In this work, we address the question of how to schedule messages for asynchronous propagation so that a fixed point is reached faster and more often. We first show that any reasonable asynchronous BP converges to a unique fixed point under conditions similar to those that guarantee convergence of synchronous BP. In addition, we show that the convergence rate of a simple round-robin schedule is at least as good as that of synchronous propagation. We then propose *residual belief propagation (RBP)*, a novel, easy-to-implement, asynchronous propagation algorithm that schedules messages in an informed way, that pushes down a bound on the distance from the fixed point. Finally, we demonstrate the superiority of RBP over state-of-the-art methods for a variety of challenging synthetic and real-life problems: RBP converges significantly more often than other methods; and it significantly reduces running time until convergence, even when other methods converge.

## 1 Introduction

Probabilistic graphical models for representing and reasoning about complex distributions have gained wide spread popularity, and are playing a role in a broad range of applications. As these models are applied to a greater variety of real-world problems, practitioners are encountering more and more networks for which inference poses a significant challenge. Consequently, the past decade has seen an explosion in the development of new methods for approximate inference in graphical models.

One of the most popular class of methods used are *message passing algorithms*, which pass messages over the graph (or a related cluster graph) until convergence. These methods, which originated with the simple *loopy belief propagation (BP)* algorithm of Pearl (1988), have been the focus of much research; multiple extensions have been proposed, and have been applied successfully to a variety of domains (e.g., (McEliece et al., 1998; Freeman and Pasztor, 2000; Taskar et al., 2004)).

Nevertheless, the application of message passing algorithms to complex, real-world networks remains problematic: BP and its extensions simply do not converge for challenging models and convergent alternatives (e.g., (Yuille, 2001; Welling and Teh, 2001)) have not been widely adopted in practice (see Section 6). Moreover, in large networks, even if convergence is possible, this may be at a significant computational cost. In practice, researchers often abandon a non-convergent model in favor of a simpler one, or simply stop the algorithm at an arbitrary point.

In this paper, we propose a very simple yet surprisingly effective method for improving the convergence properties of any message passing algorithm. Our method derives from the well-known empirical observation that *asynchronous* message passing algorithms, where messages are updated sequentially, generally converge faster and more often than the synchronous variant, where all messages are updated in parallel. Yet, many practitioners continue to use the synchronous variant, due perhaps to ease of implementation, and to the lack of clear guidelines on scheduling the messages in asynchronous propagation. When sequential updating is used, the "standard" naive schedule is one where a message is propagated as soon as one of its inputs has changed. Somewhat surprisingly, there has been virtually no attempt to study the question of determining a good order for propagation. While several scheduling variants have been considered for the special case of decoding (e.g., Wang et al. (2005)), to our knowledge, only the tree reparameterization (TRP) algorithm of Wainwright et al. (2002) proposes an asynchronous message scheduling ap-proach for the general case, and even TRP still leaves many degrees of freedom in the message scheduling order (the selection of trees and the order in which they are calibrated).

In this paper, we address the task of constructing an effective message scheduling scheme for asynchronous propagation so that convergence is achieved more often and faster. We begin by showing that any reasonable asynchronous BP converges to a unique fixed point under similar (but not the weakest) sufficient conditions to that of synchronous BP. Under these conditions, we also derive an upper bound on the convergence rate of round-robin asynchronous BP, showing that its provable convergence rate is guaranteed to be at least as good as that of synchronous BP.

Motivated by the bounds obtained in this analysis, we propose a very simple and practical *residual propagation* approach for scheduling messages in a message passing algorithm. The intuition behind residual propagation is that not all messages are equally useful towards achieving convergence. Sending a message whose current value is quite similar to its value in the previous iteration is almost redundant, while sending a message that is very different from its previous value is likely to be more informative, and lead to more rapid transfer of information throughout the network. We define the *message residual* as the magnitude of difference between two consecutive values of the message, and schedule messages in order of the largest residual. We show that this scheduling approach is a greedy algorithm for pushing down an upper bound on the distance between the current set of messages and the fixed point messages that we aim to reach; thus, the message scheduling algorithm is designed so as to try and speed up convergence.

Residual propagation is a general approach that can be applied to any problem that requires solving a set of fixed point equations. We focus on *residual belief propagation (RBP)* — its application to belief propagation. We present results for both the sum-product and max-product algorithms, applied both to challenging grid networks, and on a set of large real-life networks on which previous methods have failed. We compare both to BP with smoothing and to the TRP method (Wainwright et al., 2002), showing that RBP converges significantly more often than other methods, and in virtually all of the real-world networks. We also show that, even in convergent cases, RBP achieves convergence using far fewer messages, and significantly lower computational cost.

## 2 Propagation Based Inference

We begin by briefly reviewing the basic belief propagation algorithm. We then present it in the broader context of finding a solution to a set of fixed point equations. The remainder of our technical presentation will be formulated in this broader setting, which also encompasses a range of other inference algorithms, as well as many other problems.

Let $\mathcal{X} = \{X_1, \ldots, X_n\}$ be a finite set of random variables. We use $\mathbf{x}$ to denote an assignment to $\mathcal{X}$ and $\mathbf{x}_c$ to denote an assignment to a subset of variables $\mathbf{X}_c$. A *probabilistic graphical model* is a factored representation of a joint distribution over $\mathcal{X}$. The distribution is defined using a set of *factors* $\{\phi_c : c \in \mathcal{C}\}$, where each $c$ is associated with the variables $\mathbf{X}_c \subset \mathcal{X}$, and $\phi_c$ is a function from the set of possible assignments to $\mathbf{X}_c$ to $I\!\!R^+$. The joint distribution is defined as: $P(\mathcal{X} = \mathbf{x}) = \frac{1}{Z} \prod_{c \in \mathcal{C}} \phi_c(\mathbf{x}_c)$ where $Z$ is a normalization constant known as the *partition function*.

Message passing algorithms in a probabilistic graphical model can be defined over a special structure called a *cluster graph*. Each node $s$ in this graph corresponds to a set of cluster variables $\mathbf{X}_s$, and is associated with a factor over these variables. Clusters are connected by edges along which messages can be propagated. A message between two clusters $s$ and $t$ is a factor over the variables in their sepset $\mathbf{X}_{s,t} \subset \mathbf{X}_s \cap \mathbf{X}_t$. In *sum-product belief propagation*, this message is computed via the update equation:

$$m_{s \to t}(\mathbf{x}_{s,t}) := \sum_{\mathbf{X}_s - \mathbf{X}_{s,t}} \phi_s(\mathbf{X}_s) \prod_{r \in \mathcal{N}_s - \{t\}} m_{r \to s}(\mathbf{x}_{r,s}), \quad (1)$$

where $\mathcal{N}_s$ are all of the clusters adjacent to $s$. In principle, messages may be passed in any order. Convergence is achieved when both sides of the update equations for each cluster in the cluster graph are calibrated (equal). Indeed, Yedidia et al. (2001) provide a derivation of the convergence points defined by the belief propagation equations as fixed points of the *Bethe free energy* function.

From a more abstract perspective, we can view each message as residing in some *message space* $\mathcal{R} \subset (I\!\!R^+)^d$. Thus, we can view an entire set $\mathcal{M}$ of messages in the cluster graph as a subset of $\mathcal{R}^{|\mathcal{M}|}$. We use $m$ to index individual messages, $\mathbf{v}_m \in \mathcal{R}$ to denote the $m$'th message, and $\mathbf{v} \in \mathcal{R}^{|\mathcal{M}|}$ to denote an entire joint assignment of messages.[1] Note that in a cluster graph, we have one index $m$ for every pair of adjacent clusters $s, t$. We can now view the update rule in Eq. (1) as defining a mapping function $f_m : \mathcal{R}^{|\mathcal{M}|} \mapsto \mathcal{R}$, which defines the value of the $m$'th message as a function of (some subset of) the other messages. Our goal is to find a *fixed point* $\mathbf{z}^*$ of this set of functions — one where, for each $m$:

$$f_m(\mathbf{z}^*) = \mathbf{z}^*_m.$$

We can use these individual update functions $f_m$ to define a single global *synchronous update function* $f^s : \mathcal{R}^{|\mathcal{M}|} \mapsto \mathcal{R}^{|\mathcal{M}|}$ as follows:

$$f^s(\mathbf{v}_1, \ldots, \mathbf{v}_{|\mathcal{M}|}) = (f_1(\mathbf{v}), \ldots, f_{|\mathcal{M}|}(\mathbf{v})) \quad (2)$$

This function updates all of the coordinates simultaneously, using their previous values. We can also define a set of individual *asynchronous update functions* as follows:

$$f^a_m(\mathbf{v}_1, \ldots, \mathbf{v}_{|\mathcal{M}|}) = (\mathbf{v}_1, \ldots, f_m(\mathbf{v}), \ldots, \mathbf{v}_{|\mathcal{M}|}) \quad (3)$$

---
[1] In general, different messages reside in spaces of different dimensions, corresponding to the number of assignments in the scope of the message; for simplicity of notation, we assume that all messages have the same dimension $d$.

Synchronous propagation applies $f^s$ repeatedly, until convergence; asynchronous propagation typically applies the different $f_m^a$'s one at a time, in some order.

We note that, although we presented this abstraction in the context of the BP algorithm, it actually characterizes a very broad class of problems. Most closely related are the various variants of BP, such as max-product propagation (e.g, (Weiss and Freeman, 2001)), *generalized belief propagation (GBP)* (Yedidia et al., 2001) or *expectation propagation (EP)* (Minka, 2001); many of these variants, like sum-product BP, can actually be derived as fixed point solutions to some constrained optimization problem (as first shown by Yedidia et al. (2001)). This characterization also captures many other algorithms. For example, variational approximation methods (Jordan et al., 1998) also define the solution in terms of a set of fixed point equations, each of which defines one coordinate in terms of the others, and achieve convergence by iterated application of these equations. There are also numerous applications far outside the scope of inference in graphical models (see Section 6).

## 3 Asynchronous Belief Propagation

The availability of workstations that can perform billions of operations per second has made large scale computations practical even on a single CPU. Indeed, many fixed point computations, and practically all belief propagation runs, are carried out on a single CPU with no parallelization. In this setting, it is common wisdom that asynchronous propagation is superior to its synchronous counterpart.

Despite this, much of the theoretical work on analysis of convergence focuses on the synchronous case. In particular, to our knowledge, all theoretical guarantees regarding the convergence of belief propagation for general graphs (e.g., (Ihler et al., 2005; Mooij and Kappen, 2005)) apply only to the fully synchronous variant. In this section, we study the convergence properties of asynchronous propagation. We show that, under similar (but not the weakest) sufficient conditions to those that guarantee convergence of synchronous propagation, any reasonable asynchronous propagation is also convergent. We also analyze the convergence rate of a round-robin asynchronous algorithm, and provide bounds that are at least as good as can be provided for its asynchronous counterpart.

### 3.1 Convergence of Asynchronous Propagation

Our analysis focuses on the extent to which each application of an operator (e.g., a message passing step) reduces the distance between the current set of messages and the fixed point of the process. The basic tool used in this analysis is that of a *contraction*. Let $\mathbf{V}$ be a real, finite dimensional vector space, and let $\|\cdot\|$ denote a vector norm. A mapping $f : \mathbf{V} \to \mathbf{V}$ is a $\|\cdot\|$-*contraction* if

$$\|f(\mathbf{v}) - f(\mathbf{w})\| \leq \alpha \|\mathbf{v} - \mathbf{w}\|$$

for some $0 \leq \alpha \leq 1$, for all $\mathbf{v}, \mathbf{w} \in \mathbf{V}$. When $f$ is a contraction under some norm, we are guaranteed that it has a unique fixed point $\mathbf{z}^*$. Moreover, the sequence $f(\mathbf{v}_0), f(f(\mathbf{v})), \ldots$, where the mapping $f$ is applied repeatedly, is guaranteed to converge to $\mathbf{z}^*$ regardless of the starting point $\mathbf{v}_0$.

In order to apply this type of analysis to belief propagation, we first need to define a distance metric on the space of messages. Recall that the message space $\mathcal{R}^{|\mathcal{M}|}$ is the set of messages in the network, each of which is itself a vector in $\mathcal{R}$. Thus, the overall distance metric between two messages is an aggregate of a set of distances for individual messages. We therefore have a *message norm* $\|\mathbf{v}_m - \mathbf{w}_m\|_m$ that measures distances between individual messages, and a *global norm*, that aggregates these message distances into an overall distance metric between points in $\mathcal{R}^{|\mathcal{M}|}$. Our analysis is based on the use of the max-norm $\|\cdot\|_\infty$ for the external norm, but we take no position on the choice of the message norm. We thus define $\|\mathbf{v} - \mathbf{w}\|_\infty = \max_{m \in \mathcal{M}} \|\mathbf{v}_m - \mathbf{w}_m\|_m$.

Our analysis in this section assumes that the synchronous mapping $f^s$ (Eq. (2)) is a contraction in max-norm ($L_\infty$). Although this assumption is a fairly strong one, there are interesting conditions under which it holds. In particular, for the case of belief propagation, Mooij and Kappen (2005) give sufficient conditions for $f^s$ to be a contraction under different norms, including the max-norm. We note that Mooij and Kappen also provide weaker sufficient conditions for convergence, based on the *spectral norm* of the matrix; currently, our analysis relies on the assumption that $f^s$ is a max-norm contraction. However, even when the assumption fails to hold (as it often does), the analysis for the contractive case can shed light on cases giving rise to local convergence to a fixed point. Thus, for the rest of this section, we assume that, for any pair of points $\mathbf{v}, \mathbf{w}$ in the message space $\mathcal{R}^{|\mathcal{M}|}$, we have that $\|f^s(\mathbf{v}) - f^s(\mathbf{w})\|_\infty \leq \alpha \|\mathbf{v} - \mathbf{w}\|_\infty$. It then follows that $f^s$ has a unique fixed point $\mathbf{z}^*$, and that

$$\|f(\mathbf{v}) - \mathbf{z}^*\|_\infty \leq \alpha \|\mathbf{v} - \mathbf{z}^*\|_\infty. \tag{4}$$

We now use results developed in the field of *chaotic relaxation*, or distributed asynchronous computation of fixed points, to show that this assumption implies convergence of any reasonable asynchronous update schedule. Following the seminal paper of Chazan and Miranker (1969), we make only the following trivial assumption about the order of the updates:

**Assumption 3.1:** For every message $m$, there is a finite time $T_m$ so that for any time $t \geq 0$, the update $\mathbf{v} := f_m^a(\mathbf{v})$ is executed at least once in the time interval $[t, t + T_m]$. ∎

In other words, every message is updated infinitely often (until convergence).

**Theorem 3.2:** If $f^s$ is a max-norm contraction, then any asynchronous propagation schedule that satisfies Assumption 3.1 will converge to a unique fixed point.

This result is a direct consequence of the central theorem of Bertsekas (1983) (Section 4) and its application to the case of max-norm contractions. The intuition beyond the proof is straightforward. The key idea is that, after applying coordinate-wise operations a sufficient number of times, a point will be reached where, just as in the case of synchronous iterations, the current message will be in an $L_\infty$ sphere that is strictly confined within the sphere of previous iterations (see Bertsekas (1983, 1997) for more details).

## 3.2 Comparing the Convergence Rate of Synchronous and Asynchronous Propagation

Bertsekas (1997), in Section 6.3.5, compares the convergence rate of synchronous and asynchronous propagation in the setting of multiple CPUs and communication delays. Our setting is somewhat different: rather than (possibly arbitrary) communication delays, it is our choice of the update schedule that determines the "update time" of the inputs of messages. In synchronous propagation we are, by choice, using the input values of the previous iteration of all messages. Intuitively, we should expect to do better if more up-to-date values are used when updating a message. This intuition has wide empirical support both in applications of belief propagation and of parallel and distributed computing (see (Bertsekas, 1997) and references therein). We now show that the same methods of analysis used by Bertsekas (1997) can be used to provide a formal foundation for this intuition.

To make our analysis concrete, we consider a round-robin asynchronous message schedule; thus, at each iteration we update all messages using some predefined order $\mathbf{o}$, and the computation of a message uses the most up-to-date values of its inputs.

The global max-norm contraction of (Eq. (4)) also implies a form of local contraction. For all $m \in \mathcal{M}$, we have:

$$\|f_m(\mathbf{v}) - \mathbf{z}_m^*\|_m \leq \max_i \alpha_m^i \|\mathbf{v}_i - \mathbf{z}_i^*\|_m, \quad (5)$$

for all $\mathbf{v} \in \mathcal{R}^{|\mathcal{M}|}$. Here, $\alpha_m^i \leq \alpha$ is the local contraction factor for message $m$ relative to message $i$; this refined form allows different local contraction guarantees to hold for different messages. Using $\rho_S$ to denote the synchronous convergence rate, we then have the following upper bound

$$\|\mathbf{v}(t) - \mathbf{z}^*\|_m \leq \rho_S^t \|\mathbf{v}(0) - \mathbf{z}^*\|_\infty, \quad (6)$$

where $\rho_S = \max_{m,i} \alpha_m^i$. We now analyze the convergence rate $\rho_A$ of asynchronous updates.

**Theorem 3.3:** Let $\mathbf{o}$ be an ordering of the messages in a round-robin asynchronous iteration and let $b_\mathbf{o}^m$ be the set of messages that appear before $m$ in that order. Let $\mathbf{v}(0) \in \mathcal{R}^{|\mathcal{M}|}$ be some arbitrary starting point, and $\mathbf{v}_m(t)$ be defined via:

$$\mathbf{v}_m(t) = f_m^a(\{\mathbf{v}_i(t) : i \in b_\mathbf{o}^m\}, \{\mathbf{v}_i(t-1) : i \notin b_\mathbf{o}^m\}), \quad (7)$$

so that some of its inputs (in $b_\mathbf{o}^m$) are more up-to-date. Denoting by $\rho_m$ the message dependent convergence rate, we have that:

$$\|\mathbf{v}_m(t) - \mathbf{z}_m^*\|_m \leq \rho_m \rho_A^{t-1} \|\mathbf{v}(0) - \mathbf{z}^*\|_\infty \quad (8)$$
$$\leq \rho_A^t \|\mathbf{v}(0) - \mathbf{z}^*\|_\infty, \quad (9)$$

where $\rho_m$ is chosen to satisfy

$$\rho_m \geq \max\{\max_{i \in b_\mathbf{o}^m} \alpha_m^i \rho_i, \max_{i \notin b_\mathbf{o}^m} \alpha_m^i\}, \quad (10)$$

and $\rho_A = \max_m \rho_m$.

**Proof:** We use induction on the individual messages $\mathbf{v}_m(t)$, in the global order in which they are generated; that is, our inference proceeds simultaneously over iterations $t$ and the individual message updates within each iteration, as per Eq. (7). For all of the messages at $t = 0$, the desired result holds trivially. Now, consider an update for some message $m$ at iteration $t$. We can now write:

$$\|\mathbf{v}_m(t) - \mathbf{z}_m^*\|_m$$
$$\leq \max\left\{\max_{i \in b_\mathbf{o}^m} \alpha_m^i \|\mathbf{v}_i(t) - \mathbf{z}_i^*\|_m, \max_{i \notin b_\mathbf{o}^m} \alpha_m^i \|\mathbf{v}_i(t-1) - \mathbf{z}_i^*\|_m\right\}$$
$$\leq \max\left\{\max_{i \in b_\mathbf{o}^m} \alpha_m^i \rho_i \rho_A^{t-1}, \max_{i \notin b_\mathbf{o}^m} \alpha_m^i \rho_A^{t-1}\right\} \|\mathbf{v}(0) - \mathbf{z}^*\|_\infty$$
$$= \rho_A^{t-1} \max\left\{\max_{i \in b_\mathbf{o}^m} \alpha_m^i \rho_i, \max_{i \notin b_\mathbf{o}^m} \alpha_m^i\right\} \|\mathbf{v}(0) - \mathbf{z}^*\|_\infty$$
$$\leq \rho_A^{t-1} \rho_m \|\mathbf{v}(0) - \mathbf{z}^*\|_\infty.$$

The second line follows from Eq. (5), and the update operator defined in Eq. (7). In the third line, the first term in the brackets follows from Eq. (8) of the induction hypothesis, and the second term follows from Eq. (9). The last line follows from Eq. (10). This proves the inductive hypothesis of Eq. (8); Eq. (9) follows from the definition of $\rho^A$. ∎

Note that Eq. (10) is, in fact, a set of inequalities, one for the $\rho_m$ corresponding to each message $m$. To see that there is at least one valid solution, we set $\rho_m = \alpha$ for all $m$; as $\alpha_m^i \leq 1$, the inequality follows trivially. Indeed, if we select $\rho_A$ to be the lowest value for which Theorem 3.3 holds, it immediately follows that:

**Corollary 3.4:** *For a round-robin asynchronous iteration in some order $\mathbf{o}$ we have $\rho_A \leq \rho_S$.*

Thus, we have shown that, when max-norm contraction holds, the guarantees on convergence rate for asynchronous updates are at least as good as those for the synchronous case. But are they any better? Intuitively, it seems clear that, when some $\alpha_m^i$'s are smaller than the global $\alpha$, the convergence rate may be better. In particular, we see that $\rho_m$ is likely to be lower when $\alpha_m^i$ is lower for messages $i$ not in $b_\mathbf{o}^m$; that is, we obtain greater improvements in the convergence rate for message $m$ if its coupling to less up-to-date messages is weaker.

**Example 3.5:** To illustrate the above analysis, we consider a simple model with 4 binary variables and pairwise potentials $C_1 = \{X_1, X_2\}$, $C_2 = \{X_2, X_3\}$, $C_3 = \{X_3, X_4\}$, and $C_4 = \{X_4, X_1\}$ so that the cluster graph has a single loop with $|\mathcal{M}| = 8$ messages in all. We assign the potentials

$$\phi_1 = \begin{pmatrix} .25 & .25 \\ .5 & .25 \end{pmatrix}, \phi_2, \phi_3 = \begin{pmatrix} 1 & 0.5 \\ 0.5 & 0.5 \end{pmatrix}, \phi_4 = \begin{pmatrix} 1 & .5 \\ .5 & 1 \end{pmatrix}$$

The above model has a unique fixed point and using the analysis of Mooij and Kappen (2005) we have that the theoretical rate of contraction is $\alpha = 0.88$. We use simulation to evaluate the local contraction factors $\alpha_m^i$. We generated $500,000$ random message vectors in the 32-dimensional message space (4 values for each of the 8 messages). For each of these random vectors $\mathbf{v}$ we then computed $f_m(\mathbf{v})$ for each message $m$. We then evaluated the distance of these messages to the fixed point message vector $\mathbf{z}^*$, and compared it to the distance of the input messages. Using these distances, we estimated $\alpha_m^i[n]$ for each random sample $n$ using Eq. (5). Finally, we set $\alpha_m^i$ to be the maximum value across all random vectors in the message space. This simulation resulted in an estimated synchronous convergence rate of $\rho_S = 0.714$ which, as expected, is somewhat lower than the theoretical contraction factor. When we now solve for the individual $\rho_m$ and $\rho_A$ using Eq. (10), for some order $\mathbf{o}$, we get an asynchronous convergence rate $\rho_A$ that is often smaller than the synchronous convergence rate. Concretely, for 100 random orderings of messages, we have a mean $\rho_A = 0.678$ with a standard deviation of 0.038, demonstrating our intuition that many different message orders can provide a guaranteed convergence rate that is strictly smaller than the synchronous one. ∎

## 4 Residual Propagation

We now address the question of constructing a concrete message update schedule that achieves better convergence properties than standard synchronous or asynchronous update. Unfortunately, the analysis of the previous section does not immediately give rise to such a schedule. On the one hand, we do not, in general, know the local contraction factors $\alpha_m^i$; indeed, we want our approach to apply even in cases where the mapping $f^s$ is non-contractive, so that appropriate $\alpha_m^i$'s may not even exist. On the other hand, we do not necessarily wish to restrict our attention to a round-robin schedule. Empirically, when running BP, we see that some parts of the network converge very quickly, whereas others take much longer to reach reasonable values. As messages sent along edges where the two clusters are almost calibrated have little impact on the overall network parameterization, we are better off focusing more of our updates on the less-stable regions. Thus, we want to construct a dynamic message schedule that is based on the current state of messages rather than commit to a single round-robin ordering of messages.

Nevertheless, the analysis of Section 3.2 provides significant insight on the factors that are most important in achieving rapid convergence. As shown in the proof of Theorem 3.3, the actual bound on the distance between $\mathbf{v}_m(t)$ and its fixed point value $\mathbf{z}_m^*$ depends on the current distances $\|\mathbf{v}_i(t) - \mathbf{z}_i^*\|_m$ of its "neighboring" messages $i$. Thus, one way to speed up convergence is to choose to update the message $m$ so as to minimize the largest of these distances. Unfortunately, we cannot directly measure the distance between a current message and its unknown fixed point value. However, can provide a bound on this difference that uses easy to measure quantities

**Proposition 4.1:** *Let $\mathbf{V}$ be a real, finite dimensional vector space and $\|\cdot\|$ some vector norm over $\mathbf{V}$. Let $g$ be some mapping over $\mathbf{V}$ such that $\mathbf{z}$ is a fixed point of $g$. Then for any $\mathbf{v} \in \mathbf{V}$ and $\alpha < 1$ such that $\|g(\mathbf{v}) - \mathbf{z}\|_\infty \leq \alpha \|\mathbf{v} - \mathbf{z}\|_\infty$, we have that:*

$$\|g(\mathbf{v}) - \mathbf{z}\| \leq \|\mathbf{v} - \mathbf{z}\| - \frac{(1-\alpha)}{(1+\alpha)} \|\mathbf{v} - g(\mathbf{v})\|.$$

**Proof:** We begin by deriving, using the triangle inequality,

$$\begin{aligned} \|\mathbf{v} - g(\mathbf{v})\| &= \|\mathbf{v} - \mathbf{z} + \mathbf{z} - g(\mathbf{v})\| \\ &\leq \|\mathbf{v} - \mathbf{z}\| + \|g(\mathbf{v}) - \mathbf{z}\| \\ &\leq \|\mathbf{v} - \mathbf{z}\| + \alpha \|\mathbf{v} - \mathbf{z}\| \\ &= (1+\alpha) \|\mathbf{v} - \mathbf{z}\|, \end{aligned} \quad (11)$$

where the third line follows from the contraction property. We use contraction again to write

$$\begin{aligned} \|g(\mathbf{v}) - \mathbf{z}\| &\leq \alpha \|\mathbf{v} - \mathbf{z}\| \\ &= \|\mathbf{v} - \mathbf{z}\| - (1-\alpha) \|\mathbf{v} - \mathbf{z}\| \\ &\leq \|\mathbf{v} - \mathbf{z}\| - \frac{(1-\alpha)}{(1+\alpha)} \|\mathbf{v} - g(\mathbf{v})\|. \end{aligned}$$

∎

The above result shows that the reduction in distance between the $m$'th message and its fixed point can be bounded by some fraction (less than 1) of the difference in values of the $m$'th message before and after the update. Importantly, we note that this analysis applies at any point in the algorithm at which the update equations are a contraction mapping; there is no requirement that there be a global contraction factor $\alpha$, or even a unique fixed point to the system.

Based on this analysis, we define the *residual* for a message $m$ at the point $\mathbf{v}$ to be $r_m(\mathbf{v}) = \|f_m(\mathbf{v}) - \mathbf{v}_m\|_m$. We can now propose a simple, greedy algorithm, that aims to maximize the residual at each iteration. That is, at each step, it chooses to update the message:

$$m^t = \mathrm{argmax}_m r_m(\mathbf{v}(t)). \quad (12)$$

We note that this scheme focuses solely on the component $\|\mathbf{v}_i(t) - \mathbf{z}_i^*\|_m$ in the bound used in the proof of Theorem 3.3, completely ignoring $\alpha_m^i$. As we discussed, these contraction rates are rarely known, but if one can bound them, a more refined algorithm that took them into account would probably be better.

We also note that in sparse systems, where one message depends only on few others, the method can be implemented very efficiently: the residuals can be maintained incrementally, as the residual for a message $m$ changes only when we update a message $i$ on which $m$ depends. In fact, even when the system is not sparse, the residuals are typically maintained in any case in order to check the convergence of the algorithm. We can thus maintain a priority queue of messages to update, based on their residuals; at each step, we extract the message of highest residual from the queue, update it, and recompute the residuals of the messages that depend on it. In practice, as shown in our experiments, there is little computational cost (per update) to maintaining this data structure.

## 5 Experimental Evaluation

We set out to evaluate the effectiveness of our residual belief propagation (RBP) method along three axes: ability to converge, rate of convergence, and the quality of the marginals obtained. We compare our RBP approach to several method: Synchronous BP (SBP); Asynchronous BP (ABP) where messages are scheduled for propagation after their input has changed; The TRP method of Wainwright et al. (2002). For TRP, as the choice of spanning trees is not made concrete, we tried several variants that seem appropriate for grids including random trees, criss-cross trees, comb-like trees, and snake shaped trees. All variants performed similarly and we report results here for the snake trees (both horizontal and vertical) that were marginally better than the other TRP variants. We use standard message damping of $0.2$ for all methods (a range of values up to $0.5$ produced similar results). All algorithms use the same code base and differ only in the way messages are scheduled for propagation. Runs were performed on a Pentium 4 with 3.4GHz processor and 2GB of memory.

**Ising Grids**
We begin by considering random grids, parameterized by the Ising model. These networks provide a systematic way for evaluating an algorithm, as we can easily control both the size and difficulty of the inference task; they are also the standard benchmark for evaluating message propagation algorithms. We generate random grids with $N \times N$ binary variables as follows: A uniformly sampled univariate potential in the $[0, 1]$ range is assigned to each variable. For pairwise potentials, we use the Ising model where all edge potentials $\psi_{i,j}(X_i, X_j)$ are $e^{\lambda C}$ when $x_i = x_j$ and $e^{-\lambda C}$ otherwise. To make the inference problem challenging, we sample $\lambda$ in the range $[-0.5, 0.5]$ so that some factors reward agreement of marginal beliefs and others disagreement. Higher values of $C$ impose stronger constraints, leading to a harder inference task.

Figure 1(a) shows the cumulative percentage of convergence of the different algorithms as a function of actual CPU time, including the time required for computing the residuals and selecting the edge/tree in the RBP and TRP algorithms, respectively. By about 20 seconds, all methods reach a plateau, with minor improvements afterward (runs were allowed to continue up to 500 seconds with minor changes to the curves). Notably, RBP converges more often than all other methods and is able to converge on roughly $2/3$ of the runs for which TRP did not converge. It is also interesting to note that while TRP converges marginally faster on the relatively easy girds where convergence is rapid, RBP converges significantly faster for those grids for which TRP is slow to converge. The importance of asynchronous propagation in general is also evident as the synchronous variant is significantly inferior to to even the simple asynchronous method which is in turn inferior to both TRP and RBP.

Figure 1(b) shows the same results for harder random graphs where the difficulty parameter $C$ was increased. While all methods, as expected, converge less often, the relative benefit of RBP is greater. This phenomenon where RBP is more effective when the problem is harder was consistent across a range of grid sizes and difficulties (not shown for lack of space). It is also interesting to note that in this harder scenario TRP is only marginally superior to ABP.

We take a more global view of our results in Figure 1(c) in which we examine the number of messages propagated until convergence by TRP and RBP as a function of the number of messages propagated by ABP, a good practical measure for the difficulty of the inference task. The superiority of RBP is evident, and its advantage grows with the difficulty of the inference task.

Next, we want to address the issue of the quality of our approximation. We consider random grids of size $11 \times 11$ with $C = 11$, where exact inference was tractable, and use as our error metric the average KL-divergence between the approximate and exact node marginals. Figure 1(d) compares the quality of the fixed point of RBP vs. that of ABP (results for TRP were qualitatively the same and are not shown for clarity). For runs where both algorithm converged, both algorithms achieve a fixed point of the same quality. For runs where only RBP converged, the results are mixed, but RBP provides a better approximation overall. Note that, even in the cases where ABP has lower error than RBP, the error of RBP is low and is very close to that of ABP. For challenging networks, where the error of ABP is large, RBP is always equal to or superior to ABP. Interestingly, the results of convergent runs of RBP are not markedly worse in the cases where BP does not converge.

To demonstrate the applicability of our residual propagation scheme to other message propagation algorithms, we applied it to the max-product (MP) algorithm. Figure 2(a) shows the cumulative convergence percentage of the different methods as a function of actual CPU time for

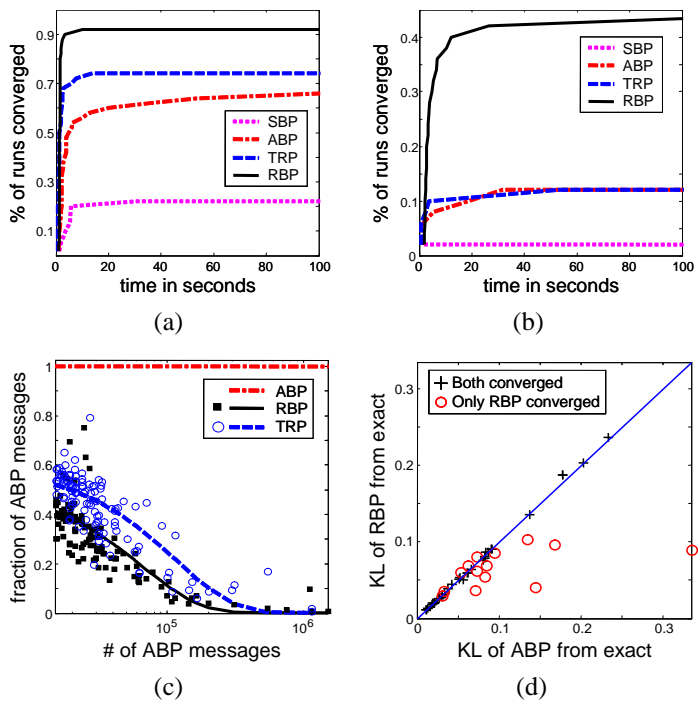

Figure 1: (a) cumulative percentage of converged runs (y-axis) as a function of actual running time (x-axis). Shown are results for SBP, ABP, TRP, and RBP for 50 random grids of size $11 \times 11$ and $C = 11$. Runs were allowed to continue for 500 seconds with marginal changes to the plot (not shown). (b) same as (a) for more difficult graphs with $C = 13$. (c) fraction of messages sent by TRP and RBP relative to the messages sent by ABP (y-axis), as a function of the number of messages sent by ABP (x-axis). Shown are a range of grids where ABP converged with sizes $7 \times 7$, $9 \times 9$, and $11 \times 11$ with $C = 7, 9, 11, 13$ (235 grids in all). The lines show an exponential fit to the points. (d) scatter plot of the average KL divergence of node beliefs from the exact node marginals of RBP (y-axis) vs. ABP (x-axis) for 50 random $11 \times 11$ grids with $C = 11$. Shown are grids where both methods converged (black '+') and grids where only RBP converged (red 'o').

50 random $7 \times 7$ grids with $C = 7$. As in the case of standard BP, our residual based scheme RMP converges significantly more often than all other methods. Interestingly, for this task which is typically recognized as more challenging than standard belief propagation (hence the use of smaller grids), the differences between the TRP based approach (TRMP) and the asynchronous (AMP) variant are not as pronounced. Thus, consistent with our previous results we see that the uninformed schedule of TRP is not sufficiently effective for more challenging inference problems. Figure 2(b) shows the same results for larger random $9 \times 9$ grids. As before, while the convergence of all methods deteriorates, the superiority of RMP over the other methods is more significant as the problem gets harder.

Finally, we also apply our method to generalized belief propagation (GBP) which is known to converge significantly more often than standard BP. We therefore focus on harder grids and compare our residual variant RGBP to GBP on $20 \times 20$ grids. Figure 2 shows that while both methods converge on all grids, our RGBP algorithm converges significantly faster. This phenomenon was consistent for $30 \times 30$ and $40 \times 40$ grids of varying difficulties (not shown for lack of space), with the advantage of RGBP growing, on average, with the difficulty of the inference task.

**Real Networks**

We now proceed to evaluating our algorithm on complex networks arising in real-world applications. We consider examples from two markedly different models in computational biology (Jaimovich et al., 2005; Yanover and Weiss, 2003). In these networks, exact inference is intractable, but BP has been shown to produce good results for smaller networks within the same general family. Thus, we can hope that BP algorithms also provide reasonable answers for larger networks.

The first domain we consider is that of predicting protein-protein interaction network from noisy genomic data. These networks, generated by Jaimovich et al. (2005), contain approximately $30,000$ binary hidden variables, corresponding to interaction relationships between pairs of proteins, and to cellular localizations of these proteins. The network is induced by a relational Markov network (Taskar et al., 2004), which defines a set of template potentials. Node potentials represent noisy observations of these variables, such as a biological assay where an interaction between two proteins was observed. There are also "triad" potentials over triples of variables, reflecting (for example) a soft constraint that two interacting proteins should be localized in the same region of the cell. These triad potentials create a large number of small loops, inducing a very difficult inference task. There are over $30,000$ potentials in the cluster graph and a similar number of loops. We consider 8 different networks with the same structure but different parameterizations (based on different learning setups in Jaimovich et al. (2005)) for which neither SBP, ABP nor TRP converged even when allowed to run for an order of magnitude longer than RBP. In contrast, RBP converged on $7/8$ networks, taking $4 - 7$ minutes to do so.

The second domain we consider is that of protein folding. Proteins have a 3D structure made up of interconnected amino acids and side-chains. Inferring this structure from the protein sequence is an important problem in computational biology. Yanover and Weiss (2003) show that inferring structure via energy minimization can be posed as an inference problem in a graphical model. The network for each protein is an independent inference task with a unique structure and parameterization, containing between hundreds and thousands of variables of cardinalities 2–81, and is highly irregular. We applied the different methods to all networks (from *www.cs.huji.ac.il/c̃heny/proteinsMRF.html*). Our implementation of ABP did not converge on 6 protein networks even when allowed to run for 30 minutes (we note that this is far fewer than the number of networks reported not to converge by Yanover and Weiss (2003)). In contrast, our RBP algorithm converged on all networks. In particular, it took an average $2\frac{1}{2}$ minutes (with a maximum of 4 minutes) to converge on those networks for which ABP did not converge. In all these models, both the synchronous SBP variant and TRP did not converge on many more networks than even ABP, again demonstrating the importance of an informed message schedule.

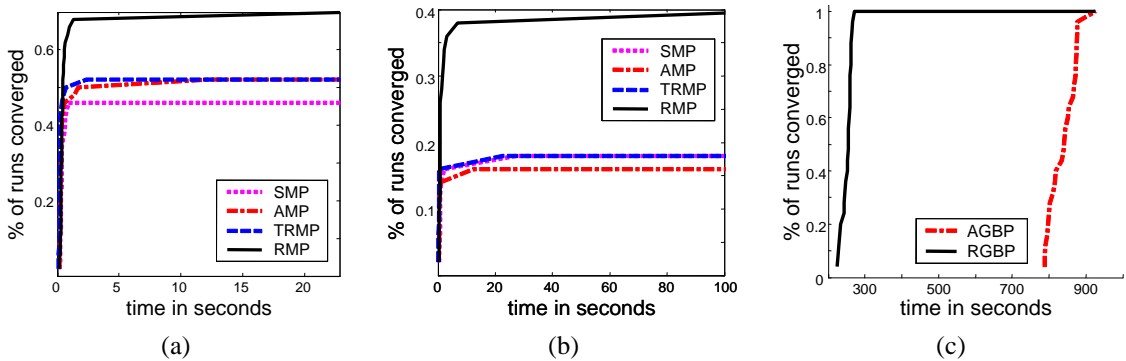

Figure 2: cumulative percentage of converged runs (y-axis) as a function of time (x-axis) for 50 random grids. (a) comparison of our RMP to the synchronous (SMP), asynchronous (AMP), and TRP (TRMP) variants of the max-product algorithm for $7 \times 7$ grids with $C = 7$. (b) same as (a) for larger $9 \times 9$ grids. (c) comparison of GBP and our RGBP method for $20 \times 20$ grids with $C = 7$.

## 6 Discussion and Future Work

In this work we addressed the task of message scheduling of propagation methods for approximate inference. We showed that any reasonable asynchronous algorithm converges under similar conditions to that of synchronous propagation and proved that the convergence rate of a round-robin asynchronous algorithm is at least as good as that of its synchronous counterpart. Motivated by this analysis, we then presented an extremely simple and efficient message scheduling approach that minimizes an upper bound on the distance of the current messages from the fixed point. We demonstrated that our algorithm is significantly superior to state-of-the-art methods on a variety of challenging synthetic and real-life problems.

Interestingly, our choice of message schedule had a significant effect not only on the rate of convergence but also on the convergence success. While this phenomenon is not typically observed in the field of decoding (see for example Kfir and Kanter (2003)), it is consistent with the observations made by Wainwright et al. (2002). We conjecture that when using more oblivious update schemes (including both synchronous and asynchronous), contradictory signals are obtained from different parts of the network, causing the oscillations commonly observed in practice. In contrast, RBP transmits information in a more "purposeful" way, potentially propagating it to other parts of the network before they have the opportunity to transmit a contradictory signal that causes oscillations.

Propagation methods that are guaranteed to converge have been proposed by Yuille (2001) and Welling and Teh (2001). These methods are fairly complex to implement; they also provide limited improvements over BP in terms of accuracy, and no improvement in convergence rate. While our methods have no convergence guarantees for general graphs, they are easy to implement, and appear to converge on almost all but very hard synthetic problems. Furthermore, our method converges much more quickly than standard BP or state-of-the-art TRP.

A number of sequential message schedules have been proposed for message decoding using belief propagation; these schedules have been shown to converge faster than synchronous updates. Some works, notably that of Wang et al. (2005), have formally analyzed convergence rates for different update schemes for low-density parity-check codes, under certain idealized assumptions, showing, for example, that a simple asynchronous propagation approach is twice as fast as the fully synchronous variant. Both the algorithms proposed in this literature and the methods used in the analysis are highly specialized to coding networks, and it is not clear how they can be applied to general inference problems outside of the field of decoding.

Our approach defines a whole family of algorithms and can be applied to practically any message propagation algorithm. We demonstrated that, in addition to improving BP, our method is effective in improving the performance of the max-product algorithm as well as that of generalized belief propagation. Importantly, our approach can in fact be applied to a wide variety of methods that iteratively apply a set of update equations until a fixed point is reached. Examples include the *information bottleneck* clustering method Tishby et al. (1999) or variational approximation methods

(e.g., Jordan et al. (1998)).

The problem of solving sets of fixed point equations arises in numerous applications far outside the realm of graphical models, including partial differential equations and solving large systems of linear equations. In such systems, the most common approaches for iterating the equations are: Jacobi, a simultaneous (synchronous) update; and Gauss-Seidel, which follows a fixed round-robin schedule. It is widely recognized that, in practice, Gauss-Seidel and related variants converge faster than Jacobi. Indeed, for the case of linear systems, there are formal results proving this fact. It is therefore somewhat surprising that the problem of intelligently scheduling the updates, whether in a round robin fashion or more generally, has been so little studied. Most results for linear systems generalizing the celebrated Stein-Rosenberg theorem (see, for example, Bertsekas (1997)), still assume a fixed cyclic order or assume that the mapping satisfies additional properties. For the case of non-linear systems, even less seems known; most analysis are for particular systems of equations and particular orderings of the updates (e.g., Porsching (1971)). The only results, to the best of our knowledge, on general asynchronous updates are focused on the case where the asynchrony results from vagaries of a parallel architecture with processing and communication delays; these results basically prove that, under certain conditions, the asynchrony does not cause too many problems, and provide conditions under which favorable architecture of communication delays may even improve the convergence rate. Further improvements are shown for special cases such as that of monotone contraction mappings (e.g., Tsitsiklis (1989)). Other than the results on Gauss-Seidel method and its variants, there is no analysis that attempts to *design* asynchrony into the system so as to achieve better convergence. Thus, it would be intriguing to consider the application of residual propagation to the broad range of tasks that are naturally modeled as the solution to a set of fixed point equations.

## Acknowledgments

We thank Gene Golub and Dimitri Bertsekas for useful pointers. This work was supported by the DARPA transfer learning program under contract FA8750-05-2-0249.